\documentclass[letterpaper, 10 pt, conference]{ieeeconf}  

\IEEEoverridecommandlockouts                              

\usepackage{amsmath} 
\usepackage{amssymb}  
\usepackage{graphicx} 
\usepackage{booktabs}

\usepackage{algorithm}
\usepackage{algpseudocode}
\usepackage[font=footnotesize]{caption}
\usepackage[subrefformat=parens]{subcaption}
\usepackage{stfloats}

\newtheorem{lemma}{Lemma}
\newenvironment{proofsketch}
{\noindent\textit{Proof Sketch:}}
{\hfill$\square$}

\usepackage{threeparttable}

\usepackage{hyperref}

\title{\LARGE \bf
VGFM: Expressive Robot Policies via Dense Value Guidance in Flow Matching
}

\author{Prajwal Koirala$^{1}$ and Mark Campbell$^{1}$
\thanks{*This work was supported by NSF CPS grant CNS-2211599 and NSF FRR grant IIS-2305532.}
\thanks{$^{1}$Prajwal Koirala and Mark Campbell are with the Sibley School of Mechanical and Aerospace Engineering, Cornell University.
        {\tt\small \{pk596, mc288\}@cornell.edu}}%
\thanks{Code is available at: \url{https://github.com/PrajwalKoirala/VGFM_Policy} .}
}

\begin{document}

\maketitle
\thispagestyle{empty}
\pagestyle{empty}

\begin{abstract}
Recent robot learning paradigms increasingly rely on large offline datasets of robotic interactions to train control policies. Expressive generative models enable rich and multimodal action representations, expanding the capability of this paradigm for complex robotic control. However, policy improvement with multi-step generative actors remains challenging. In offline reinforcement learning (RL), incorporating value-based objectives along generative trajectories often introduces substantial training complexity, including backpropagation through time (BPTT), auxiliary architectures, or distillation losses. We propose Value-Guided Flow Matching (VGFM), a scalable offline RL framework that enables dense value-guided shaping within a flow-based policy while avoiding BPTT and additional algorithmic overhead. VGFM parameterizes the policy as a conditional flow-matching model in action (x-prediction) space, ensuring that each intermediate flow step produces a valid robot action that can be directly evaluated by a standard offline RL critic. This design allows value guidance to be applied at randomly sampled flow times without differentiating through the entire generative trajectory, while preserving inference-time flexibility by varying the discretization of the underlying flow ODE without retraining. Evaluated on robotic locomotion and manipulation tasks in OGBench, VGFM achieves strong performance across a wide range of tasks under rigorous evaluation protocols. With minimal hyperparameter tuning, these results demonstrate that VGFM provides a simple, scalable, and effective approach for expressive policy learning in long-horizon, goal-oriented robotic control.
\end{abstract}


\section{Introduction}
Learning from pre-collected static robotic datasets has become a key paradigm for training control policies without additional environment interaction. By leveraging large-scale demonstrations or logged trajectories, robots can acquire complex skills while avoiding the safety risks and high costs associated with online exploration. Offline reinforcement learning (RL) provides a principled framework for this setting, allowing agents to optimize policies solely from previously collected data without requiring further interaction with the environment. A central challenge in offline RL is distribution shift: the learned robot policy may produce actions that fall outside the dataset, leading to unsafe or ineffective behaviors \cite{levine2020offline,chi2023diffusion}. As robotic datasets become more diverse and heterogeneous, the underlying state–action distributions often exhibit strong multimodality and compositional structure, demanding expressive policy classes that can accurately represent such complexity. This motivates the use of generative models to construct expressive, yet behaviorally consistent policies \cite{chi2023diffusion}.

Recent advances in generative models (like diffusion \cite{sohl2015deep, ho2020denoising} and flow-matching \cite{liu2022flow, lipman2022flow}) have demonstrated strong capacity to represent complex, multimodal conditional distributions, making them appealing for expressive policy learning in offline reinforcement learning (RL). In realistic domains such as robotics and autonomous driving, offline datasets typically aggregate diverse, task-relevant strategies—such as varying grasp orientations, approach angles, or conflicting topological choices at road intersections (e.g., turning left versus continuing straight)—that are poorly captured by unimodal policy parameterizations. By enabling policies to faithfully model this behavioral diversity while remaining tightly constrained to the data distribution, generative policies offer a principled mechanism for improving generalization and robustness in offline RL, while maintaining strong behavioral constraints from data \cite{wang2022diffusion, chi2023diffusion, zheng2025diffusion}.

\begin{figure}[t]
    \centering
    \begin{subfigure}{\linewidth}
        \centering
        \includegraphics[width=\linewidth]{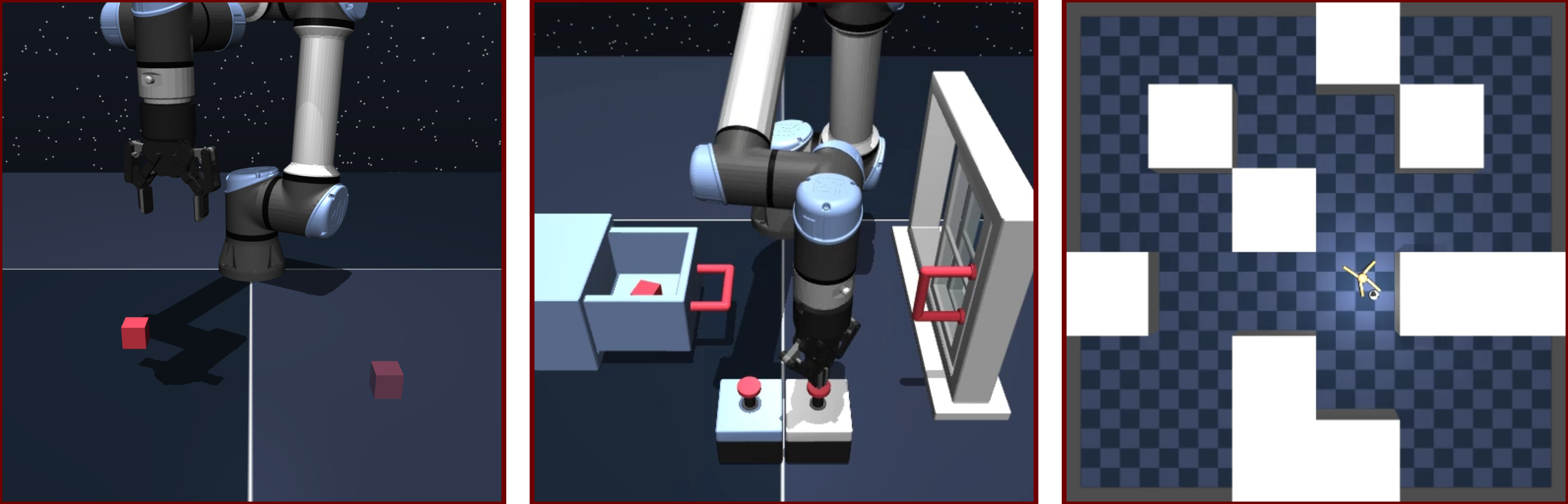}
        \vspace{-15pt}
        \caption{OGBench tasks}
        \label{fig:ogb_tasks}
        \vspace{15pt}
    \end{subfigure}
    \begin{subfigure}{\linewidth}
        \centering
        \includegraphics[width=\linewidth]{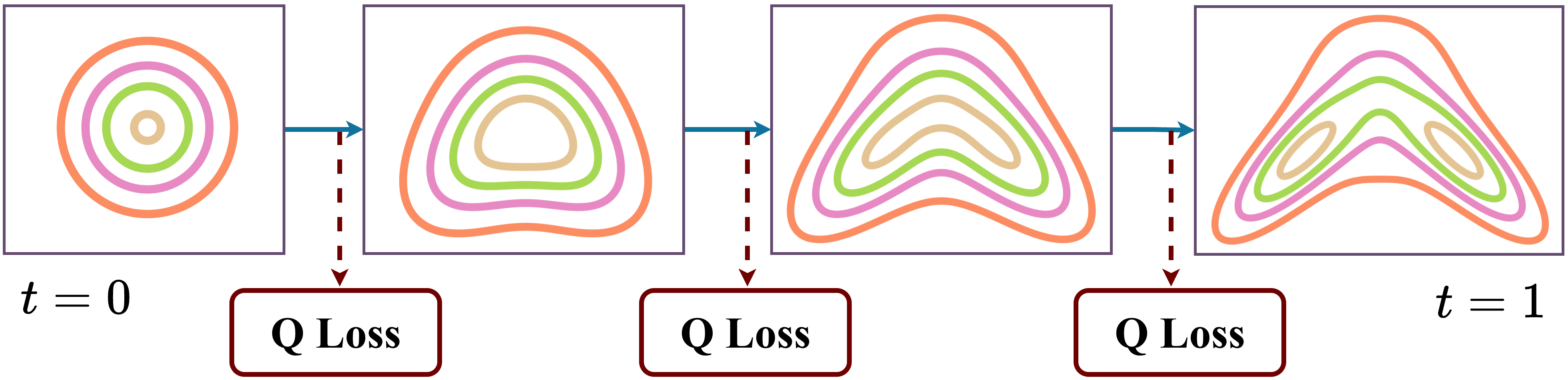}
        \caption{VGFM framework}
        \label{fig:vgfm_intro}
    \end{subfigure}
    \vspace{-10pt}
    \caption{Overall illustration of the proposed VGFM framework and benchmark tasks it is evaluated on. (a) Examples of robotic manipulation and locomotion environments used for evaluation, spanning maze navigation, goal-directed interaction, and multi-stage object manipulation. (b) Illustration of how VGFM applies value guidance (Q-loss) to each intermediate step of a flow-based policy. A 2D unimodal Gaussian is transported to a multimodal 2D action distribution over multiple steps, with Q-loss applied at each intermediate flow step.}
    \label{fig:ogb_tasks_and_vgfm}
    \vspace{-10pt}
\end{figure}

However, integrating diffusion- and flow-based policies into value-based offline reinforcement learning remains nontrivial. In standard off-policy actor–critic frameworks such as DDPG+BC (or TD3+BC \cite{fujimoto2019off}), policy improvement requires computing gradients of the action-value function with respect to the policy’s output action. When the policy is parameterized as a flow-matching model, this action is generated via a multi-step continuous-time generative process, typically instantiated through numerical ODE integration of a learned velocity field. Consequently, propagating the policy improvement signal necessitates backpropagation through the entire flow trajectory. This reliance on backpropagation through time (BPTT) incurs substantial computational overhead and can introduce numerical instability, particularly as the number of integration steps increases or when deploying expressive, high-dimensional policy parameterizations \cite{park2025flow}.

Several recent works attempt to address this limitation. For example, Flow Q-Learning (FQL) \cite{park2025flow} avoids BPTT by distilling the learned generative policy into a separate single-step network, simplifying training but sacrificing inference-time scalability, as multi-step generation is no longer available. Scalable Offline RL (SORL) \cite{espinosa2025scaling} introduces shortcut models that allow inference-time trade-offs between accuracy and computation, but still requires BPTT during training through the generative process. Single-Step Completion Policies (SSCP) \cite{koirala2025flow} avoid BPTT by predicting completion vectors within a single network, but empirical results suggest limited gains from increased inference-time computation, restricting the benefits of multi-step generation.

In this work, we propose a simple and scalable alternative. We apply a standard DDPG+BC-style policy improvement objective directly to a flow-matching policy by predicting end-target actions at each flow step, while applying learning signals in the velocity (or displacement) space. This decoupling enables each flow step to be supervised by both behavioral cloning and value-based policy improvement without requiring backpropagation through the generative trajectory. As a result, the policy can be discretized arbitrarily and supports inference-time scalability—allowing multi-step generation when compute is available—while maintaining stable and efficient training. Empirically, we show that this value-guided flow formulation scales to expressive, multi-step policies and achieves strong performance on standard offline RL benchmarks, particularly in settings where multimodality and action diversity are essential.

Key contributions of this work are as follows.
\begin{itemize}
    \item We introduce a value-guided formulation of flow matching policies, interpreting each step of the flow as a return-optimizing decision in action space, and applying a DDPG+BC-style objective at every step to enable value-guided policy improvement.
    \item We introduce an end-target prediction objective with velocity-space losses that enables policy improvement without backpropagation through the generative trajectory in an offline RL setting.
    \item We demonstrate that this approach supports inference-time scalability with multi-step generation and achieves strong empirical performance in most robotic manipulation and locomotion tasks in OGBench \cite{park2024ogbench} benchmark.
\end{itemize}

\section{Preliminaries}
\subsection{Offline Reinforcement Learning}

We consider a Markov Decision Process (MDP) $\mathcal{M} = (\mathcal{S}, \mathcal{A}, p, r, \rho, \gamma)$,
where $\mathcal{S}$ and $\mathcal{A}$ denote the state and action spaces,
$p(s'|s, a)$ is the transition dynamics,
$r(s, a)$ is the reward function,
$\rho(s)$ is the initial state distribution,
and $\gamma \in [0,1)$ is the discount factor.
A stochastic policy $\pi_\theta(a | s): \mathcal{S} \rightarrow \Delta(\mathcal{A})$ induces the trajectory distribution
$p^{\pi_\theta}(\tau) = \rho(s_0)\prod_{h=0}^{T-1} \pi_\theta(a_h \mid s_h)p(s_{h+1}\mid s_h,a_h)$,
and the expected return is
$J(\pi_\theta) = \mathbb{E}_{\tau \sim p^{\pi\theta}} \left[\sum_{h=0}^T \gamma^h r(s_h, a_h)\right]$.

In the offline RL setting, the agent has access only to a fixed dataset
$\mathcal{D} = {(s, a, r, s')}$ collected by an unknown behavior policy $\pi_\beta$,
without further environment interaction.
A key challenge in this setting is distributional shift:
when $\pi_\theta$ selects out-of-distribution (OOD) actions not covered by $\mathcal{D}$,
the corresponding value estimates $Q(s, a)$ become unreliable,
leading to performance degradation~\cite{fujimoto2019off, kumar2020conservative}.
To mitigate this, many offline RL algorithms regularize the learned policy to remain close to the behavior distribution,
often via a divergence constraint:
\begin{equation}\label{eq:offline_constraint}
\max_{\pi}  J(\pi)
\quad \text{s.t. } D_{\mathrm{KL}}(\pi(\cdot \mid s), \pi_\beta(\cdot \mid s)) \leq \delta.
\end{equation}
A practical instantiation of this principle is the behavior-regularized actor-critic framework,
which augments the actor loss with a behavior cloning (BC) term ~\cite{fujimoto2021minimalist, wu2019behavior}:
\begin{equation}\label{eq:brac_actor}
\mathcal{L}\pi(\theta)
= \mathbb{E}{(s,a)\sim\mathcal{D}}
\big[-Q_\phi(s, \pi_\theta(s)) + \alpha \log \pi_\theta(a \mid s)\big],
\end{equation}
where $\alpha > 0$ balances exploitation and conservatism.
This formulation enables stable policy optimization by constraining updates to remain within the data support.
In most existing methods, $\pi_\theta$ is modeled as a unimodal Gaussian,
which simplifies gradient computation but limits expressiveness in multimodal datasets.

\subsection{Diffusion and Flow-Based Generative Models}

\paragraph{Diffusion models.}
Diffusion models~\cite{sohl2015deep, ho2020denoising, song2020score}
learn to reverse a gradual noising process that transforms data into noise through a stochastic differential equation (SDE):
\begin{equation}
d\mathbf{x}_t = f(t)\mathbf{x}_t dt + g(t) d\mathbf{w}_t,
\end{equation}
where $\mathbf{w}_t$ is standard Brownian motion and $f(t), g(t)$ specify the drift and diffusion schedules.
A neural network is trained to approximate the reverse-time dynamics, typically by predicting the noise added at each step via denoising score matching.
Sampling then proceeds by numerically integrating a reverse-time SDE or its deterministic counterpart, the probability flow ODE.
While diffusion models achieve high-quality generative performance, they often require many iterative steps at inference time,
which limits their practicality for control applications.

\paragraph{Flow matching.}
Flow matching~\cite{lipman2022flow, liu2022flow} provides a deterministic alternative to diffusion-based generative modeling by directly learning a time-dependent velocity field that transports a simple base distribution to the data distribution.
Let $p_0(\mathbf{z}) = \mathcal{N}(\mathbf{0}, I)$ denote the base distribution and $p_1(\mathbf{x}) = p_{\text{data}}(\mathbf{x})$ the target data distribution.
Intermediate states are constructed via linear interpolation between base and data samples:
\begin{equation}
\mathbf{x}_t = (1 - t)\mathbf{z} + t\mathbf{x}, \quad t \sim \mathcal{U}[0,1].
\end{equation}
Flow matching learns a velocity field $\mathbf{v}_\theta(\mathbf{x}_t, t)$ that matches the optimal velocity vector field $\mathbf{x} - \mathbf{z}$ along these interpolated paths by minimizing
\begin{equation}
\mathcal{L}_{\text{FM}}(\theta)
= \mathbb{E}_{t, \mathbf{z}, \mathbf{x}}
\big[ \lVert \mathbf{v}_\theta(\mathbf{x}_t, t) - (\mathbf{x} - \mathbf{z}) \rVert_2^2 \big],
\end{equation}
where $\mathbf{z} \sim p_0$ and $\mathbf{x} \sim p_{\text{data}}$.
At inference time, new samples are generated by integrating the learned ordinary differential equation
\begin{equation}
\frac{d\mathbf{x}_t}{dt} = \mathbf{v}_\theta(\mathbf{x}_t, t),
\quad \mathbf{x}_0 \sim \mathcal{N}(\mathbf{0}, I),
\end{equation}
which deterministically maps latent samples to the data manifold.

\section{Method: Value-Guided Flow Matching (VGFM)}
\label{sec:combined_method}

\subsection{Problem Formulation and Motivation}
\label{subsec:problem}

We build upon the KL-regularized policy improvement objective commonly adopted in offline RL~\cite{wu2019behavior, peng2019advantage}:
\begin{equation}
\begin{aligned}
\pi^\star
=
\arg\max_{\pi}
\;
\mathbb{E}_{s \sim \mathcal{D}}
\Big[ &
\mathbb{E}_{a \sim \pi(\cdot|s)}[Q(s,a)]
- \\
&\alpha \, \mathrm{KL}\big(\pi(\cdot|s)\,\|\,\pi_\beta(\cdot|s)\big)
\Big]
\end{aligned}
\label{eq:kl_obj}
\end{equation}
where $Q(s,a)$ is a critic trained via standard offline RL procedures (e.g., TD3+BC style) and $\alpha > 0$ controls the strength of behavior regularization. The solution to Eq.~\eqref{eq:kl_obj} admits the advantage-weighted form
\begin{equation}
\label{eq:awp}
\pi^\star(a|s)
\propto
\pi_\beta(a|s)\,
\exp\!\left(\tfrac{1}{\alpha} Q(s,a)\right),
\end{equation}
and taking gradients with respect to the action yields the score decomposition
\begin{equation}
\label{eq:score_decomp}
\nabla_a \log \pi^\star(a|s)
=
\nabla_a \log \pi_\beta(a|s)
+
\tfrac{1}{\alpha}\,
\nabla_a Q(s,a).
\end{equation}
This decomposition reveals policy improvement as the superposition of two gradients: behavior matching ($\nabla_a \log \pi_\beta$) and value guidance ($\nabla_a Q$).

Eq.~\eqref{eq:score_decomp} implies that value gradients can guide policy improvement at every intermediate point along a generative trajectory, rather than only at the terminal action, enabling dense credit assignment across denoising steps. However, realizing this in flow- or diffusion-based policies is challenging. Conventional approaches such as DQL \cite{wang2022diffusion} and CAC \cite{ding2023consistency} require propagating value gradients through the entire generative process, which entails backpropagation through time (BPTT) over many discretized steps. For long horizons, BPTT becomes memory-intensive and computationally costly, and in offline RL it can introduce instability due to vanishing/exploding gradients and compounded approximation errors. To address these limitations, prior work either distills the generative policy into a separate network~\cite{park2025flow, chen2023score} or introduces auxiliary objectives (e.g., shortcut or completion losses) to reduce the effective generation depth~\cite{koirala2025flow, espinosa2025scaling}. Some recent methods like \cite{ki2025actor} also use a time-conditioned noise-level critic for dense guidance along the generative path.

To retain dense value guidance without incurring full BPTT, we introduce a flow-matching parameterization whose 
x-prediction formulation \cite{li2025back} enables BPTT-free value guidance. Importantly, consistent with policy-constrained offline RL, we do not require globally accurate critics; instead, we assume that the critic provides locally meaningful gradients within dataset-supported regions of the action space. Our objective is not to compute the exact optimizer of Eq.~\eqref{eq:kl_obj}, but to design an inference-time scalable flow-based policy that leverages value gradients densely along the generative trajectory while preserving behavior regularization.

\subsection{Flow-Based Policy Parameterization}
\label{subsec:flow}

To realize the decomposition in Eq.~\eqref{eq:score_decomp} within a generative policy framework, we parameterize the conditional policy $\pi_\theta(a \mid s)$ as a flow-matching model that deterministically transports samples from a base distribution $p_0 = \mathcal{N}(\mathbf{0}, I)$ to actions in the dataset-supported region. For a given state $s$, we define a continuous trajectory $\{a_t\}_{t \in [0,1]} \subset \mathbb{R}^d$ via linear interpolation between a latent sample $a_0 \sim p_0$ and a terminal action $a_1$:
\begin{equation}
\label{eq:interpolation}
a_t = (1 - t) a_0 + t a_1, \quad t \in [0,1].
\end{equation}
The generative process is governed by an ordinary differential equation
\begin{equation}
\label{eq:ode}
\frac{d a_t}{d t} = u_\theta(s, a_t, t),
\end{equation}
where $u_\theta: \mathcal{S} \times \mathbb{R}^d \times [0,1] \to \mathbb{R}^d$ is a neural velocity field. Integrating Eq.~\eqref{eq:ode} from $t=0$ to $t=1$ with initial condition $a_0 \sim p_0$ yields a stochastic action sample $a_1$, with policy expressivity controlled by the expressiveness of $u_\theta$ rather than by distributional assumptions (e.g., Gaussian nature of $p_0$). 

\paragraph{Flow matching as behavior regularization.}
Rather than training the flow via likelihood maximization or adversarial objectives, we adopt flow matching~\cite{lipman2022flow} to directly enforce proximity to the empirical behavior distribution. Given a state-action pair $(s, a_1) \sim \mathcal{D}$ and a base sample $a_0 \sim p_0$, we construct the interpolated point $a_t$ via Eq.~\eqref{eq:interpolation} for $t \sim \mathcal{U}[0,1]$ and minimize the regression objective
\begin{equation}
\label{eq:fm_loss}
\mathcal{L}_{\mathrm{FM}}(\theta)
=
\mathbb{E}_{\substack{(s,a_1) \sim \mathcal{D} \\ a_0 \sim p_0 \\ t \sim \mathcal{U}[0,1]}}
\Big[
\big\|
u_\theta(s, a_t, t) - (a_1 - a_0)
\big\|_2^2
\Big].
\end{equation}
Critically, $\mathcal{L}_{\mathrm{FM}}$ implements behavior regularization \emph{by construction}: the optimal velocity field transports mass from $p_0$ precisely toward the empirical action distribution in $\mathcal{D}$ induced by $\pi_\beta$. Unlike KL penalties added post-hoc to policy gradients, flow matching embeds conservatism directly into the policy parameterization, ensuring that even intermediate flow states remain anchored to dataset-supported regions.

\paragraph{x-prediction as the enabler of BPTT-free guidance}
We implement the velocity field via an \emph{x-prediction} parameterization~\cite{li2025back}, where the network directly predicts the terminal action $\hat{a}_\theta(s, a_t, t)$ and the velocity is recovered analytically:
\begin{equation}
\label{eq:xpred}
u_\theta(s, a_t, t)
=
\frac{\hat{a}_\theta(s, a_t, t) - a_t}{1 - t}.
\end{equation}
While x-prediction, together with the v-loss (eg. in Eq. \eqref{eq:fm_loss}), is motivated by performance and stability in general generative task in \cite{li2025back}, we emphasize its \emph{conceptual significance} in this work for value-guided policy learning:

\begin{lemma}[Action-Space Validity]
\label{lem:action_validity}
Under the x-prediction parameterization in Eq.~\eqref{eq:xpred}, for any state $s$, intermediate flow state $a_t$, and time $t \in [0,1)$, the prediction $\hat{a}_\theta(s, a_t, t)$ lies in the action space $\mathcal{A}$ and can be directly evaluated by a time-independent critic $Q_\phi(s, a)$.
\end{lemma}
\begin{proofsketch}
By construction, $\hat{a}_\theta(s, a_t, t)$ is a direct prediction in the action space rather than a velocity or noise vector. The flow-matching objective in Eq.~\eqref{eq:fm_loss} further ensures that, in expectation over $t$, these predictions remain regularized toward the empirical action distribution: minimizing $\|u_\theta - (a_1 - a_0)\|^2$ enforces consistency between $\hat{a}_\theta(s, a_t, t)$ and the dataset action $a_1$ across all interpolation times. Consequently, $\hat{a}_\theta(s, a_t, t)$ remains within regions where $Q_\phi$ provides meaningful gradients.
\end{proofsketch}

This property, action-space validity with implicit behavior regularization, is the key enabler for dense value guidance without backpropagation through time. As we show next, it allows us to apply value-based supervision at arbitrary flow times using a standard behavior-regularized off-policy policy gradient method, bypassing the need for noise-level critics or flow trajectory-level differentiation.

\subsection{Value Guidance Without Backpropagation Through Time}
\label{subsec:value_guidance}

Section~\ref{subsec:problem} identified a fundamental tension: while the score decomposition in Eq.~\eqref{eq:score_decomp} suggests that value gradients could guide policy improvement at every point along a generative trajectory, prior implementations require backpropagation through time (BPTT) over the flow traejctory. Lemma~\ref{lem:action_validity} resolves this tension by guaranteeing that, under x-prediction with flow matching, intermediate predictions $\hat{a}_\theta(s,a_t,t)$ remain valid actions within dataset-supported regions. This property enables a simple yet powerful alternative: applying value-based supervision \emph{directly to intermediate predictions} using a standard, time-independent critic—without differentiating through the generative trajectory.

\paragraph{Dense value supervision via action-space predictions}
For a state $s \sim \mathcal{D}$, dataset action $a_1 \sim \mathcal{D}$, base sample $a_0 \sim p_0$, and time $t \sim \mathcal{U}[0,1]$, we construct the interpolated state $a_t$ via Eq.~\eqref{eq:interpolation} and apply value guidance to the predicted terminal action $\hat{a}_\theta(s,a_t,t)$:
\begin{equation}
\label{eq:q_loss}
\mathcal{L}_{\pi_{\mathrm{Q}}}(\theta)
=
-
\mathbb{E}_{\substack{(s,a_1) \sim \mathcal{D} \\ a_0 \sim p_0 \\ t \sim \mathcal{U}[0,1]}}
\Big[
Q_\phi\big(s, \hat{a}_\theta(s,a_t,t)\big)
\Big].
\end{equation}
Critically, gradients of $\mathcal{L}_{\mathrm{Q}}$ flow \emph{only through the network parameters at the sampled time $t$}, not through the flow dynamics or across multiple time steps. Each training iteration thus provides an independent supervision signal that encourages the velocity field to steer intermediate flow states toward higher-value terminal actions, realizing the value-guidance term $\nabla_a Q(s,a)$ in Eq.~\eqref{eq:score_decomp} \emph{densely along the flow trajectory} without BPTT.

\paragraph{Combined objective and interpretation}
We train the flow matching policy by jointly optimizing behavior regularization (Eq. \ref{eq:fm_loss}) and value guidance (Eq. \ref{eq:q_loss}):
\begin{equation}
\label{eq:full_loss}
\mathcal{L}(\theta)
=
\alpha \, \mathcal{L}_{\mathrm{FM}}(\theta)
+
 \mathcal{L}_{\pi_{\mathrm{Q}}}(\theta),
\end{equation}
where $\alpha \geq 0$ controls the relative strength of regularization. This objective implements a practical approximation to the KL-regularized policy improvement in Eq.~\eqref{eq:kl_obj}, where:
\begin{itemize}
    \item $\mathcal{L}_{\mathrm{FM}}$ enforces the behavior-matching term $\nabla_a \log \pi_\beta(a|s)$ by anchoring the entire flow trajectory to the empirical action distribution.
    \item $\mathcal{L}_{\mathrm{Q}}$ implements the value-guidance term $\nabla_a Q(s,a)$ at \emph{every sampled flow time} $t$, providing dense supervision without generative trajectory differentiation.
\end{itemize}
The result is a policy that remains conservative with respect to the dataset while being biased toward higher-value actions \emph{throughout} the generative process—not merely at the terminal step.

Overall, training follows an actor–critic paradigm, analogous to TD3+BC, with the combined policy objective assuming a structure similar to Eq.~\eqref{eq:brac_actor}. The critic is optimized concurrently using the standard Bellman backup loss:
\begin{equation} \label{eq:critic_loss}
\begin{aligned}
\mathcal{L}_Q(\phi)
= &\mathbb{E}_{(s, a, r, s') \sim \mathcal{D}}
\Bigg[
\Big(
Q_\phi(s, a)
- \\
&\Big(
r + \gamma \cdot \min_{i=1,2}
Q_{\phi'_i}(s', \pi_\theta(s'))
\Big)
\Big)^2
\Bigg].
\end{aligned}
\end{equation}

\subsection{Algorithmic Properties and Relation to Prior Work}
\label{subsec:properties}

\begin{algorithm}[!h]
\caption{VGFM Training}
\label{alg:vgfm}
\begin{algorithmic}[1]
\Require Dataset $\mathcal{D}$, critic $Q_\phi$, 
x-pred network $a_\theta$ (inducing velocity field $u_\theta$ and 
flow-matching policy $\pi_\theta$)
\For{$N$ training iterations}
    \State Sample batch $\{(s, a, r, s')\} \sim \mathcal{D}$
    
    \State \textit{// Critic $Q_\phi$ update Eq. \eqref{eq:critic_loss}}
    \State Update $\phi$ using clipped double Q-learning
    
    \State \textit{// Flow policy $\pi_\theta$ update Eq. \eqref{eq:full_loss}}
    \State $a_0 \sim \mathcal{N}(\mathbf{0},\mathbf{I})$, $a_1 \gets a$, $t \sim \mathcal{U}[0,1)$
    \State $a_t \gets (1-t)a_0 + t a_1$ 
    \State $\hat{a}_\theta \gets \hat{a}_\theta(s, a_t, t)$
    \State $u_\theta \gets (\hat{a}_\theta - a_t)/(1-t)$
    \State  $\mathcal{L}_{\mathrm{FM}} \gets \|u_\theta - (a_1 - a_0)\|^2$ 
    \State $\mathcal{L}_{\mathrm{Q}} \gets -Q_\phi(s, \hat{a}_\theta)$
    \State $\theta \gets \theta - \eta_\theta \nabla_\theta \big( \alpha \mathcal{L}_{\mathrm{FM}} + \mathcal{L}_{\mathrm{Q}} \big)$ 
    
    \State Update target networks via Polyak averaging
\EndFor
\State \Return $\theta$
\end{algorithmic}
\end{algorithm}

\begin{algorithm}[!h]
\caption{Inference with VGFM Policy $\pi_\theta$}
\label{alg:inference_vgfm}
\begin{algorithmic}[1]
\Procedure{$\pi_\theta$}{$s$} \Comment{Induced policy via $\hat{a}_\theta$}
    \State \quad $x_0 \sim \mathcal{N}(0, I)$, \quad $\tau \gets 0$
    \For{T steps}
    \State $\hat{u} \gets (\hat{a}_\theta(s, x_0, \tau)-x_0)/(1-\tau)$
    \State $x_0 \gets x_0 + \hat{u}/T$
    \State $\tau \gets \tau + 1/T$
    \EndFor
    \State \textbf{return} \quad $x_0$
\EndProcedure
\end{algorithmic}
\end{algorithm}

Training, as detailed in Algorithm \ref{alg:vgfm}, proceeds by iterative critic updates and policy updates using the objectives in Eq.~\eqref{eq:critic_loss} and Eq.~\eqref{eq:full_loss} respectively. At each policy update, we sample states and actions from $\mathcal{D}$, base samples from $p_0$, and flow times uniformly from $[0,1]$, then jointly minimize $\mathcal{L}_{\mathrm{FM}}$ and $\mathcal{L}_{\mathrm{Q}}$ via standard gradient descent. At inference time, actions are generated by numerically integrating the learned ODE (Eq.~\ref{eq:ode}) from $t=0$ to $t=1$ with methods like Euler's method; the number of integration steps can be chosen to trade computation for performance without retraining. The inference procedure is summarized in Algorithm \ref{alg:inference_vgfm}.

\begin{table}[!h]
\caption{
Comparison of generative policy methods along three practical dimensions.
}
\label{tab:comparison}
\centering
\begin{tabular}{lccc}
\toprule
Method & BPTT & Inference Scalability & Auxiliary Loss \\
\midrule
DQL~\cite{wang2022diffusion}    & Yes  & Yes & None \\
FQL~\cite{park2025flow}    & No   & No  & Distillation Loss \\
SSCP~\cite{koirala2025flow}  & No   & Poor  & Completion Loss \\
SORL~\cite{espinosa2025scaling}& Yes  & Yes & Shortcut Loss \\
\textbf{VGFM (Ours)}          & \textbf{No}  & \textbf{Yes} & \textbf{None} \\
\bottomrule
\end{tabular}
\end{table}

Table~\ref{tab:comparison} positions Value-Guided Flow Matching (VGFM) against representative flow-based offline RL methods along three axes critical to practical deployment: BPTT requirement, inference flexibility, and algorithmic simplicity. Prior approaches exhibit inherent tradeoffs: methods avoiding BPTT (eg. FQL, SSCP) sacrifice inference-time scalability by requiring policy distillation or fixed-step completion vectors; methods supporting scalable inference (DQL, SORL) rely on backpropagation through time, introducing associated training complexities; and most require auxiliary losses beyond standard policy optimization. VGFM uniquely combines BPTT-free training, inference-time step adaptability, and minimal auxiliary machinery, requiring only the flow-matching and value-guidance terms already present in the core objective.

\begin{figure*}[t]
    \centering
    \begin{subfigure}{.24\textwidth}
        \centering
        \includegraphics[width=\linewidth]{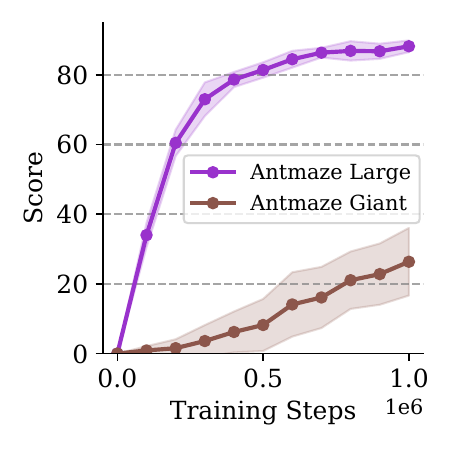}
        \caption{Antmaze Large and Giant}
        \label{fig:antmaze_large_giant}
    \end{subfigure}
    \begin{subfigure}{.24\textwidth}
        \centering
        \includegraphics[width=\linewidth]{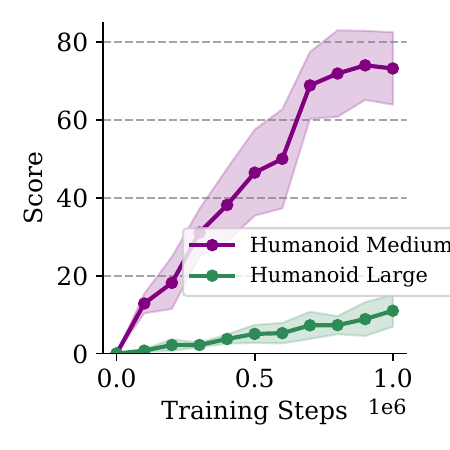}
        \caption{Humanoid Medium and Large}
        \label{fig:humanoidmaze_medium_large}
    \end{subfigure}
    \begin{subfigure}{.24\textwidth}
        \centering
        \includegraphics[width=\linewidth]{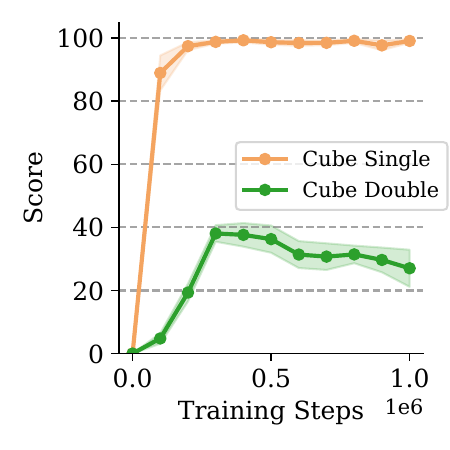}
        \caption{Cube Single and Double}
        \label{fig:cube_single_double}
    \end{subfigure}
    \begin{subfigure}{.24\textwidth}
        \centering
        \includegraphics[width=\linewidth]{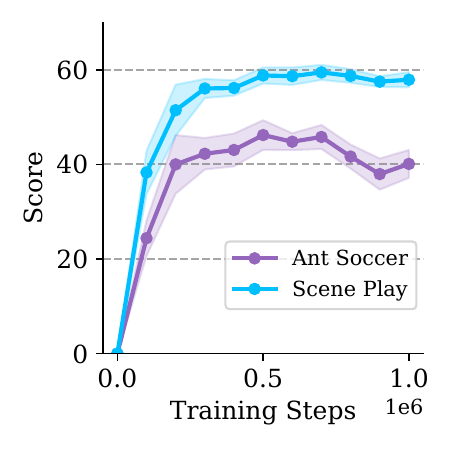}
        \caption{Ant Soccer and Scene Play}
        \label{fig:ant_soccer_scene}
    \end{subfigure}
    \caption{Training curves of VGFM for all 8 OGBench task groups. Plots show normalized return versus gradient steps. Solid lines denote mean performance over 8 random training seeds; shaded regions indicate $\pm$ one standard deviation.}
    \label{fig:full_grid_traning_curve}
\end{figure*}

\begin{table*}[b]
\vspace{10pt}
\caption{
\footnotesize
\textbf{OGBench Offline RL results.}
For each task group, we highlight the highest-performing score with boldface and additionally bold any method whose performance is within $95\%$ of the best. VGFM achieves the best or near-best performance on most of the OGBench robotic locomotion and manipulation tasks. 
}
\label{table:offline_rl_ogbench_results}
\centering
\scalebox{0.99}
{
\begin{threeparttable}
\begin{tabular}{lcccccccccccc}
\toprule
\multicolumn{1}{c}{} & \multicolumn{3}{c}{\texttt{Gaussian Policies}} & \multicolumn{6}{c}{\texttt{Generative Policies}} \\
\cmidrule(lr){2-4} \cmidrule(lr){5-10}
\texttt{Task Category} & \texttt{BC} & \texttt{IQL} & \texttt{ReBRAC} & \texttt{IDQL} & \texttt{CAC} & \texttt{IFQL} & \texttt{FQL} & \texttt{SORL} & \texttt{VGFM} \\
\midrule
\texttt{antmaze-large-singletask ($\mathbf{5}$ tasks)} & $11$ {\tiny $\pm 1$} & $53$ {\tiny $\pm 3$} & 81 {\tiny $\pm 5$} & $21$ {\tiny $\pm 5$} & $33$ {\tiny $\pm 4$} & $28$ {\tiny $\pm 5$} & 79 {\tiny $\pm 3$} & $\mathbf{89}$ {\tiny $\pm 2$} & $\mathbf{87}$ {\tiny $\pm 1$} \\
\texttt{antmaze-giant-singletask ($\mathbf{5}$ tasks)} & $0$ {\tiny $\pm 0$} & $4$ {\tiny $\pm 1$} & $\mathbf{26}$ {\tiny $\pm 8$} & $0$ {\tiny $\pm 0$} & $0$ {\tiny $\pm 0$} & $3$ {\tiny $\pm 2$} & $9$ {\tiny $\pm 6$} & $9$ {\tiny $\pm 6$} & $\mathbf{23}$ {\tiny $\pm 9$}  \\
\texttt{humanoidmaze-medium-singletask ($\mathbf{5}$ tasks)} & $2$ {\tiny $\pm 1$} & $33$ {\tiny $\pm 2$} & $22$ {\tiny $\pm 8$} & $1$ {\tiny $\pm 0$} & $53$ {\tiny $\pm 8$} & $60$ {\tiny $\pm 14$} & $58$ {\tiny $\pm 5$} & $64$ {\tiny $\pm 4$} & $\mathbf{73}$ {\tiny $\pm 8$} \\
\texttt{humanoidmaze-large-singletask ($\mathbf{5}$ tasks)} & $1$ {\tiny $\pm 0$} & $2$ {\tiny $\pm 1$} & $2$ {\tiny $\pm 1$} & $1$ {\tiny $\pm 0$} & $0$ {\tiny $\pm 0$} & $\mathbf{11}$ {\tiny $\pm 2$} & $4$ {\tiny $\pm 2$} & $5$ {\tiny $\pm 2$} & $\mathbf{9}$ {\tiny $\pm 3$} \\
\texttt{antsoccer-arena-singletask ($\mathbf{5}$ tasks)} & $1$ {\tiny $\pm 0$} & $8$ {\tiny $\pm 2$} & $0$ {\tiny $\pm 0$} & $12$ {\tiny $\pm 4$} & $2$ {\tiny $\pm 4$} & $33$ {\tiny $\pm 6$} & $60$ {\tiny $\pm 2$} & $\mathbf{69}$ {\tiny $\pm 2$} & $39$ {\tiny $\pm 2$}  \\
\texttt{cube-single-singletask ($\mathbf{5}$ tasks)} & $5$ {\tiny $\pm 1$} & $83$ {\tiny $\pm 3$} & $91$ {\tiny $\pm 2$} & $\mathbf{95}$ {\tiny $\pm 2$} & $85$ {\tiny $\pm 9$} & $79$ {\tiny $\pm 2$} & $\mathbf{96}$ {\tiny $\pm 1$} & $\mathbf{97}$ {\tiny $\pm 1$} & $\mathbf{99}$ {\tiny $\pm 1$} \\
\texttt{cube-double-singletask ($\mathbf{5}$ tasks)} & $2$ {\tiny $\pm 1$} & $7$ {\tiny $\pm 1$} & $12$ {\tiny $\pm 1$} & $15$ {\tiny $\pm 6$} & $6$ {\tiny $\pm 2$} & $14$ {\tiny $\pm 3$} & $\mathbf{29}$ {\tiny $\pm 2$} & $25$ \tiny{$\pm 3$} & $\mathbf{29}$ {\tiny $\pm 3$} \\
\texttt{scene-singletask ($\mathbf{5}$ tasks)} & $5$ {\tiny $\pm 1$} & $28$ {\tiny $\pm 1$} & $41$ {\tiny $\pm 3$} & $46$ {\tiny $\pm 3$} & $40$ {\tiny $\pm 7$} & $30$ {\tiny $\pm 3$} & $\mathbf{56}$ {\tiny $\pm 2$} & $\mathbf{57}$ {\tiny $\pm 2$} & $\mathbf{58}$ {\tiny $\pm 2$}  \\
\bottomrule
\end{tabular}
\end{threeparttable}
}
\end{table*}

\section{Results and Discussion}

\subsection{Experimental Setup}

\subsubsection{Task Details}
The selected environments span a diverse set of challenging robotic control problems in OGBench, covering both locomotion and manipulation, and all use flat state representations. We adopt the goal-oriented sparse-reward single-task variants for all experiments. As described in the official implementation \cite{park2025flow, park2024ogbench}, each environment provides five distinct tasks, denoted by suffixes \texttt{singletask-task1} through \texttt{-task5}, corresponding to fixed evaluation goals; we train and evaluate on all five tasks per simulation environment (or task group). The locomotion suite includes AntMaze and HumanoidMaze, which require navigation of quadrupedal (8 DOF) and humanoid (21 DOF) agents through complex maze layouts, as well as AntSoccer, which extends the Ant agent’s locomotion to goal-directed ball interaction. The manipulation suite includes Cube and Scene, both involving multi-DOF robotic arms: Cube focuses on placing cube-shaped objects into specified goal configurations, while Scene requires sequencing multiple subtasks (up to eight per episode). The datasets provide semi-sparse reward annotations based on the number of remaining subtasks in manipulation environments and based on goal attainment in locomotion environments.

\subsubsection{Evaluation Protocol}
We evaluate VGFM against eight competitive baselines on the locomotion and manipulation benchmarks in OGBench, strictly adhering to the experimental protocol of \cite{park2025flow, park2024ogbench}. Following prior work, we consider the reward-based single-task variants and train each policy for $1$M gradient steps, with evaluations every $100$K steps using $50$ rollouts per checkpoint. For reporting, we adopt the official OGBench evaluation scheme and compute the average success rate over the final three checkpoints (800K, 900K, and 1M). Consequently, each table entry aggregates results over last $3$ evaluation epochs $\times$ $50$ episode rollouts $\times$ $8$ random seeds $\times$ $5$ tasks, yielding $6{,}000$ runs per task per method in Table~\ref{table:offline_rl_ogbench_results}. Although results are presented as eight task groups 
, each group comprises five distinct single-task environments; consistent with standard benchmarking practice, we train a separate policy per task, resulting in $40$ independently trained policies overall. 

\subsubsection{Baselines}
We benchmark VGFM against a representative spectrum of offline RL methods spanning Gaussian, diffusion, and flow-based policy classes. The Gaussian baselines include \emph{Behavioral Cloning} (BC)~\cite{pomerleau1988alvinn}, \emph{Implicit Q-Learning} (IQL)~\cite{kostrikov2021offline}, and \emph{ReBRAC}~\cite{tarasov2023revisiting}. Diffusion-based methods comprise \emph{Implicit Diffusion Q-Learning} (IDQL)~\cite{hansen2023idql}, and \emph{Consistency Actor-Critic} (CAC)~\cite{ding2023consistency}. Finally, we consider flow-based approaches, including \emph{Implicit Flow Q-Learning} (IFQL)~\cite{hansen2023idql,park2025flow}, \emph{Flow Q-Learning} (FQL)~\cite{park2025flow}, and SORL~\cite{espinosa2025scaling}. Results for all baselines except SORL are taken from \cite{park2025flow}, which provides extensively tuned and standardized OGBench evaluations; SORL results are from the original publication.  This consolidated setup enables a rigorous and carefully controlled empirical comparison across modeling paradigms.

\subsection{OGBench Results}
Under this statistically rigorous protocol, VGFM achieves the best or near-best performance on the majority of task groups in OGBench suite, as summarized in Table~\ref{table:offline_rl_ogbench_results}. Across this standardized large-scale benchmark spanning forty different robotic manipulation and locomotion tasks, VGFM consistently matches or surpasses established offline RL baselines, demonstrating strong performance across diverse task families. Fig.~\ref{fig:full_grid_traning_curve} further presents the full training curves for all eight task groups, reporting normalized return (y-axis) versus gradient steps (x-axis). Curves show the mean over $8$ random seeds, with shaded regions indicating $\pm$ one standard deviation. Beyond strong final performance, the training curves illustrate generally consistent improvement trends over training steps and stable learning dynamics across task groups. Collectively, these results support VGFM as an effective general-purpose approach for learning expressive policies in long-horizon robotic tasks.

\subsubsection{Hyperparameters}
We fix all architectural and optimization settings across tasks and tune only a single coefficient, $\alpha$, in Eq.~\eqref{eq:full_loss} (also see Algorithm~\ref{alg:vgfm}). This coefficient balances the policy optimization objective and behavior cloning regularization, similar in spirit to DDPG+BC-style methods, where the relative weighting between Q-loss and BC loss must be carefully tuned. In contrast to several recent approaches that introduce additional auxiliary objectives (e.g., distillation or shortcut losses), which might often require extra weighting coefficients, VGFM relies solely on a single parameter, $\alpha$, thereby avoiding additional hyperparameter interactions and increased training complexity. Moreover, inference-time compute can be adjusted via the discretization step count $T$ (Sec.~\ref{sec:inference_scalability}) without retraining.

Key hyperparameters are shared across all tasks: both actor and critic use trunk networks of size $(512,512,512,512)$, with learning rate $3\times10^{-4}$ and batch size $256$. The only task-specific tuning concerns the coefficient $\alpha$ in Eq.~\eqref{eq:full_loss}. For the results reported in Table~\ref{table:offline_rl_ogbench_results}, we use $\alpha=5$ for \texttt{Ant} tasks (all \texttt{large}, \texttt{giant}, and \texttt{AntSoccer}), $\alpha=10$ for \texttt{HumanoidMaze} (both \texttt{medium} and \texttt{large}), $\alpha=500$ for \texttt{Cube-Single}, $\alpha=50$ for \texttt{Cube-Double}, and $\alpha=50$ for \texttt{Scene-Play}.

\begin{figure}[t]
    \centering
    \includegraphics[width=0.99\linewidth]{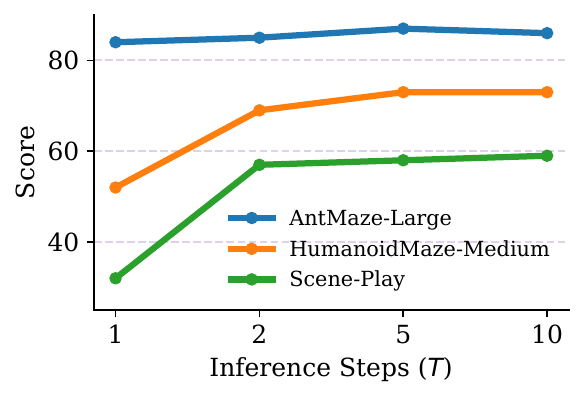}
    \caption{Task success vs. Euler steps $T$ during policy inference. Multi-step flow integration improves control precision and long-horizon task performance in OGBench robotics environments, positioning VGFM as an expressive inference-time scalable method.}
    \label{fig:inf_scalability-result}
\end{figure}

Figure~\ref{fig:inf_scalability-result} analyzes inference-time scalability in VGFM by varying the number of Euler integration steps used to solve the flow ODE during policy inference. Specifically, we vary the number of discretization steps $T \in \{1,2,5,10\}$ (from a total range of 1 to 10) in Algorithm~\ref{alg:inference_vgfm}, corresponding to progressively finer discretizations of the interval $[0,1]$. Importantly, the same trained model is used for all values of $T$ without retraining, demonstrating purely test-time scaling of compute.

\subsubsection{Inference-time Scalability} \label{sec:inference_scalability}

\begin{figure}[!h]
    \centering
    \includegraphics[width=\linewidth]{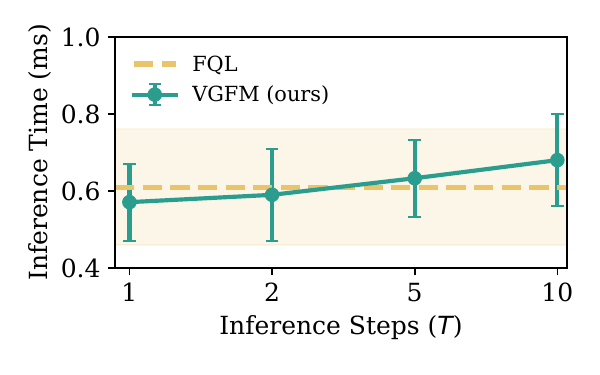}
    \caption{Inference cost vs. discretization level $T$. Despite improved performance at larger inference steps $T$, VGFM incurs near-constant wall-clock cost, comparable to single-step FQL (all evaluations done on an NVIDIA RTX A6000 GPU).}
    \label{fig:inference_time_comparison}
\end{figure}

We evaluate this effect across three representative OGBench task groups. In most environments, increasing the number of inference steps improves the average task success rate, indicating that additional test-time compute enhances policy expressivity and control precision. This demonstrates a favorable compute–performance trade-off for our flow-based generative policies, as multi-step inference requires multiple function evaluations. In practice, however, the wall-clock overhead is not proportional to the required additional number of evaluations as Euler steps increases. Instead, the average inference time is comparable to the single-step generation baseline FQL, as shown in Fig.~\ref{fig:inference_time_comparison}. Thus, finer flow discretization yields consistent performance gains while incurring only marginal additional runtime in our implementation, enabling practical test-time scaling.

As a practical compromise between inference cost and performance, we adopt $T=5$ as the default inference setting (except for \texttt{Cube-Single}, where we use $T=2$). This choice corresponds to the results presented in table \ref{table:offline_rl_ogbench_results}, and also aligns with recent findings emphasizing the role of expressive multi-step inference in OGBench-style long-horizon tasks \cite{espinosa2025scaling}. Notably, while several recent multi-step generative policy methods also demonstrate inference-time scalability in offline RL, many require backpropagation through time (BPTT) during training. In contrast, VGFM achieves scalable inference without BPTT, as summarized in Table~\ref{tab:comparison}, which compares generative policy methods along three practical dimensions: BPTT requirement, inference scalability, and reliance on auxiliary policy losses. VGFM offers a simple yet effective design across these axes.

\section{Conclusion}
In this work, we present VGFM, a flow-based generative policy framework for offline reinforcement learning. Across a rigorous evaluation on OGBench, VGFM achieves competitive or state-of-the-art performance on diverse locomotion and manipulation tasks under a standardized large-scale protocol. Beyond strong aggregate results, VGFM supports inference-time scalability through adjustable ODE discretization, enabling test-time compute–performance trade-offs without retraining. The method maintains minimal training complexity by tuning only a single task-specific coefficient and avoids BPTT, distinguishing it from several recent multi-step generative approaches. Overall, VGFM provides a simple, scalable, and effective framework for learning expressive policies in long-horizon robotic control.

\addtolength{\textheight}{-12cm}   








\bibliographystyle{IEEEtran}
\bibliography{IEEEabrv,bib}

\begin{thebibliography}{10}
\providecommand{\url}[1]{#1}
\csname url@samestyle\endcsname
\providecommand{\newblock}{\relax}
\providecommand{\bibinfo}[2]{#2}
\providecommand{\BIBentrySTDinterwordspacing}{\spaceskip=0pt\relax}
\providecommand{\BIBentryALTinterwordstretchfactor}{4}
\providecommand{\BIBentryALTinterwordspacing}{\spaceskip=\fontdimen2\font plus
\BIBentryALTinterwordstretchfactor\fontdimen3\font minus \fontdimen4\font\relax}
\providecommand{\BIBforeignlanguage}[2]{{%
\expandafter\ifx\csname l@#1\endcsname\relax
\typeout{** WARNING: IEEEtran.bst: No hyphenation pattern has been}%
\typeout{** loaded for the language `#1'. Using the pattern for}%
\typeout{** the default language instead.}%
\else
\language=\csname l@#1\endcsname
\fi
#2}}
\providecommand{\BIBdecl}{\relax}
\BIBdecl

\bibitem{levine2020offline}
S.~Levine, A.~Kumar, G.~Tucker, and J.~Fu, ``Offline reinforcement learning: Tutorial, review, and perspectives on open problems,'' \emph{arXiv preprint arXiv:2005.01643}, 2020.

\bibitem{chi2023diffusion}
C.~Chi, Z.~Xu, S.~Feng, E.~Cousineau, Y.~Du, B.~Burchfiel, R.~Tedrake, and S.~Song, ``Diffusion policy: Visuomotor policy learning via action diffusion,'' \emph{arXiv preprint arXiv:2303.04137}, 2023.

\bibitem{sohl2015deep}
J.~Sohl-Dickstein, E.~Weiss, N.~Maheswaranathan, and S.~Ganguli, ``Deep unsupervised learning using nonequilibrium thermodynamics,'' in \emph{International conference on machine learning}.\hskip 1em plus 0.5em minus 0.4em\relax pmlr, 2015, pp. 2256--2265.

\bibitem{ho2020denoising}
J.~Ho, A.~Jain, and P.~Abbeel, ``Denoising diffusion probabilistic models,'' \emph{Advances in neural information processing systems}, vol.~33, pp. 6840--6851, 2020.

\bibitem{liu2022flow}
X.~Liu, C.~Gong, and Q.~Liu, ``Flow straight and fast: Learning to generate and transfer data with rectified flow,'' \emph{arXiv preprint arXiv:2209.03003}, 2022.

\bibitem{lipman2022flow}
Y.~Lipman, R.~T. Chen, H.~Ben-Hamu, M.~Nickel, and M.~Le, ``Flow matching for generative modeling,'' \emph{arXiv preprint arXiv:2210.02747}, 2022.

\bibitem{wang2022diffusion}
Z.~Wang, J.~J. Hunt, and M.~Zhou, ``Diffusion policies as an expressive policy class for offline reinforcement learning,'' \emph{arXiv preprint arXiv:2208.06193}, 2022.

\bibitem{zheng2025diffusion}
Y.~Zheng, R.~Liang, K.~Zheng, J.~Zheng, L.~Mao, J.~Li, W.~Gu, R.~Ai, S.~E. Li, X.~Zhan \emph{et~al.}, ``Diffusion-based planning for autonomous driving with flexible guidance,'' \emph{arXiv preprint arXiv:2501.15564}, 2025.

\bibitem{fujimoto2019off}
S.~Fujimoto, D.~Meger, and D.~Precup, ``Off-policy deep reinforcement learning without exploration,'' in \emph{International conference on machine learning}.\hskip 1em plus 0.5em minus 0.4em\relax PMLR, 2019, pp. 2052--2062.

\bibitem{park2025flow}
S.~Park, Q.~Li, and S.~Levine, ``Flow q-learning,'' \emph{arXiv preprint arXiv:2502.02538}, 2025.

\bibitem{espinosa2025scaling}
N.~Espinosa-Dice, Y.~Zhang, Y.~Chen, B.~Guo, O.~Oertell, G.~Swamy, K.~Brantley, and W.~Sun, ``Scaling offline rl via efficient and expressive shortcut models,'' \emph{arXiv preprint arXiv:2505.22866}, 2025.

\bibitem{koirala2025flow}
P.~Koirala and C.~Fleming, ``Flow-based single-step completion for efficient and expressive policy learning,'' \emph{arXiv preprint arXiv:2506.21427}, 2025.

\bibitem{park2024ogbench}
S.~Park, K.~Frans, B.~Eysenbach, and S.~Levine, ``Ogbench: Benchmarking offline goal-conditioned rl,'' \emph{arXiv preprint arXiv:2410.20092}, 2024.

\bibitem{kumar2020conservative}
A.~Kumar, A.~Zhou, G.~Tucker, and S.~Levine, ``Conservative q-learning for offline reinforcement learning,'' \emph{Advances in neural information processing systems}, vol.~33, pp. 1179--1191, 2020.

\bibitem{fujimoto2021minimalist}
S.~Fujimoto and S.~S. Gu, ``A minimalist approach to offline reinforcement learning,'' \emph{Advances in neural information processing systems}, vol.~34, pp. 20\,132--20\,145, 2021.

\bibitem{wu2019behavior}
Y.~Wu, G.~Tucker, and O.~Nachum, ``Behavior regularized offline reinforcement learning,'' \emph{arXiv preprint arXiv:1911.11361}, 2019.

\bibitem{song2020score}
Y.~Song, J.~Sohl-Dickstein, D.~P. Kingma, A.~Kumar, S.~Ermon, and B.~Poole, ``Score-based generative modeling through stochastic differential equations,'' \emph{arXiv preprint arXiv:2011.13456}, 2020.

\bibitem{peng2019advantage}
X.~B. Peng, A.~Kumar, G.~Zhang, and S.~Levine, ``Advantage-weighted regression: Simple and scalable off-policy reinforcement learning,'' \emph{arXiv preprint arXiv:1910.00177}, 2019.

\bibitem{ding2023consistency}
Z.~Ding and C.~Jin, ``Consistency models as a rich and efficient policy class for reinforcement learning,'' \emph{arXiv preprint arXiv:2309.16984}, 2023.

\bibitem{chen2023score}
H.~Chen, C.~Lu, Z.~Wang, H.~Su, and J.~Zhu, ``Score regularized policy optimization through diffusion behavior,'' \emph{arXiv preprint arXiv:2310.07297}, 2023.

\bibitem{ki2025actor}
D.~Ki, H.-J. Ahn, K.~Kim, and B.-J. Lee, ``Actor-critic without actor,'' \emph{arXiv preprint arXiv:2509.21022}, 2025.

\bibitem{li2025back}
T.~Li and K.~He, ``Back to basics: Let denoising generative models denoise,'' \emph{arXiv preprint arXiv:2511.13720}, 2025.

\bibitem{pomerleau1988alvinn}
D.~A. Pomerleau, ``Alvinn: An autonomous land vehicle in a neural network,'' \emph{Advances in neural information processing systems}, vol.~1, 1988.

\bibitem{kostrikov2021offline}
I.~Kostrikov, A.~Nair, and S.~Levine, ``Offline reinforcement learning with implicit q-learning,'' \emph{arXiv preprint arXiv:2110.06169}, 2021.

\bibitem{tarasov2023revisiting}
D.~Tarasov, V.~Kurenkov, A.~Nikulin, and S.~Kolesnikov, ``Revisiting the minimalist approach to offline reinforcement learning,'' \emph{Advances in Neural Information Processing Systems}, vol.~36, pp. 11\,592--11\,620, 2023.

\bibitem{hansen2023idql}
P.~Hansen-Estruch, I.~Kostrikov, M.~Janner, J.~G. Kuba, and S.~Levine, ``Idql: Implicit q-learning as an actor-critic method with diffusion policies,'' \emph{arXiv preprint arXiv:2304.10573}, 2023.

\end{thebibliography}

\end{document}